\documentclass[10pt,twocolumn,letterpaper]{article}

\usepackage{cvpr}              

\usepackage{graphicx}
\usepackage{amsmath}
\usepackage{amssymb}
\usepackage{booktabs}
\usepackage{appendix}
\usepackage{animate}

\usepackage[pagebackref,breaklinks,colorlinks]{hyperref}

\usepackage[capitalize]{cleveref}
\crefname{appendix}{Appendix}{Appendices}
\Crefname{appendix}{Appendix}{Appendices}
\crefname{section}{Sec.}{Secs.}
\Crefname{section}{Section}{Sections}
\Crefname{table}{Table}{Tables}
\crefname{table}{Tab.}{Tabs.}

\def\cvprPaperID{11572} 
\def\confName{CVPR}
\def\confYear{2022}

\usepackage{xcolor}

\usepackage{hyperref}       
\usepackage{url}            
\usepackage{booktabs}       
\usepackage{amsfonts}       
\usepackage{nicefrac}       
\usepackage{microtype}      
\usepackage{amsmath}
\usepackage{amssymb}
\usepackage{amsthm}
\usepackage{lipsum}
\usepackage{mathtools}

\usepackage{braket}
\usepackage{algorithm}
\usepackage{algpseudocode}
\usepackage{multirow}
\usepackage{bbm}

\def \Pr {\mathrm{Pr}}

\def \saliency {\textup{\saliency}}

\def \path {\mathit{path}}

\DeclarePairedDelimiter\norm{\lVert}{\rVert}

\DeclareMathOperator*{\argmax}{arg\,max}

\DeclareMathOperator{\matrixtranspose}{T}
\newcommand{\T}{{\matrixtranspose}}

\newcommand{\bfI}{{\mathbf{I}}}

\newcommand{\bfp}{{\mathbf{p}}}

\newcommand{\bfr}{{\mathbf{r}}}

\newcommand{\bft}{{\mathbf{t}}}

\newcommand{\bfx}{{\mathbf{x}}}

\newcommand{\bfz}{{\mathbf{z}}}

\newcommand{\bftheta}{{\boldsymbol \theta}}

\newcommand{\bfmu}{{\boldsymbol \mu}}

\newcommand{\bfpi}{{\boldsymbol \pi}}

\newcommand{\bfsigma}{{\boldsymbol \sigma}}

\newcommand{\bfphi}{{\boldsymbol \phi}}

\newcommand{\bfomega}{{\boldsymbol \omega}}

\begin{document}

\title{Probabilistic PolarGMM: Unsupervised Cluster Learning of Very Noisy Projection Images of Unknown Pose}

\author{
  Supawit Chockchowwat \\
  The University of Texas at Austin \\
  {\tt\small chockchowwatsc@utexas.edu} \\
  \and
  Chandrajit L. Bajaj \\
  The University of Texas at Austin \\
  {\tt\small bajaj@cs.utexas.edu} \\
}
\maketitle

\begin{abstract}
    A crucial step in single particle analysis (SPA) of cryogenic electron microscopy (Cryo-EM), 2D classification and alignment takes a collection of noisy particle images to infer orientations and group similar images together. Averaging these aligned and clustered noisy images produces a set of clean images, ready for further analysis such as 3D reconstruction. Fourier-Bessel steerable principal component analysis (FBsPCA) enables an efficient, adaptable, low-rank rotation operator. We extend the FBsPCA to additionally handle translations. In this extended FBsPCA representation, we use a probabilistic polar-coordinate Gaussian mixture model to learn soft clusters in an unsupervised fashion using an expectation maximization (EM) algorithm. The obtained rotational clusters are thus additionally robust to the presence of pairwise alignment imperfections. Multiple benchmarks from simulated Cryo-EM datasets show probabilistic PolarGMM's improved performance in comparisons with standard single-particle Cryo-EM tools, EMAN2 and RELION, in terms of various clustering metrics and alignment errors.
\end{abstract}

\section{Introduction}





Cryo-EM captures a micrograph of small specimens frozen in thin vitreous ice via high-voltage electron beam. A micrograph consists of many particle images in random orientations. Due to particle's fragility and apparatus' imperfection, these Cryo-EM particle images are inevitably too noisy to grasp structural details with naive 3D reconstruction algorithms. A common pipeline for single particle reconstruction (SPR) includes image filtering, CTF correction, particle picking (boxing), 2D clustering and averaging, 3D reconstruction, backprojection and further refinements, generally in this order. Our work focuses on the unsupervised 2D clustering problem, namely, given a set of 2D Cryo-EM particle projection images, we group these images into multiple distinct clusters with a translation and rotational  alignment for each image such that the average of aligned images in each cluster reflects an improved signal-to-noise ratio projection of the particle from each specific orientation.

At its core, 2D translation and planar rotation clustering relies on uncorrelated noise and a highly correlated signal across well-aligned images; hence, averages would then produce a higher signal-to-noise ratio (SNR) than raw particle images. Such factored (translation and rotation) alignments and clustering problems are non-trivial, despite considering only rigid alignments SE(2), i.e. a combination of an image rotation and 2D translation, for 3 continuous degrees of freedom, per image. Multiplied by the number of images, the dimensionality of the alignment problem is simply intractable for a naive black-box optimizer. On top of the translation alignment, orientation clustering is also difficult. The number of possible orientations are infinite, diversified by the precision of possible projections with small but significant differences in projection angles. The promise of 3D reconstruction is that with a sufficiently large number of different aligned clusters, one hopes to cover all 3D particle orientations, of course with  each cluster having sufficient density or number of intra-cluster (similar) particle projection images. To add to the challenge, many biological specimens have multiple (usually unknown) 3D non-crystalline symmetries, making many distinct 3D orientations indistinguishable in their projections. There are also cases where the specimens have preferred orientations and hence violating the 3D recovery via uniform distribution of SO(3) orientations. For example, cylindrical archaeal 20S proteasome naturally aligns itself with its axis or sides facing the ice surface. While not in the scope of this paper, it is worth noting that single particle Cryo-EM  projection images can contain additional irregularities such as specimen inhomogeneity, structural conformation and deformation changes of the 3D particles, thereby violating the single 3D particle assumption for the projected images.





Based on FBsPCA, we present a probabilistic polar-coordinate learning of a Gaussian mixture model (PolarGMM) to solve the aforementioned 2D particle clustering problem. FBsPCA provides an efficient image rotation operator leading naturally to soft assignment mixture modeling optimized by the expectation maximization (EM) algorithm. To handle translation alignment, an EM centering estimate  approximates the center of each particle image. We then incorporate translation to FBsPCA, producing 2 product operators, each having different computational costs at pre-processing and post-clustering time. Evaluated with multiple  datasets, PolarGMM clusters and aligns particle images more accurately than standard SPR tools such as EMAN2 and RELION do, with a comparable alignment.
\section{Related Works}

In practice, Cryo-EM users usually rely on single particle analysis toolkits which implement their different particle clustering and averaging algorithms. For example, EMAN2 utilizes reference-free methods as its initial particle projection clustering and averaging. It finds translational and rotational invariant features through a self-correlation function (SCF), and computes auto-correlations on each polar-transformed rings \cite{corrfn}. It then performs $k$-means clustering in multivariate statistical analysis (MSA) bases. This approach sometimes fails due to the severe speckled noise in the projection images, and so EMAN2 relies on the ability to recover a correct clustering via many iterations of 3D backprojections and re-clustering. Another example is the approach taken by RELION's 2D particle clustering (ML2D). The ML2D optimizes a log likelihood function via expectation maximization algorithm (EM) \cite{relionmain} \cite{SCHERES2005139} marginalized over the single parameter of in-plane rotations \cite{relionexplain}. IMAGIC \cite{imagic2d} performs MSA for an efficient hierarchical classification which builds reference sets for multi-reference alignment (MRA) class averaging. Xmipp's clustering 2D (CL2D) classifies and aligns each sample to a class average, and carefully attempts to balancing the cluster size by splitting any dominating super-cluster \cite{SORZANO2010197}. The methods in this article while having some resemblance to all these prior methods, tradeoff adaptable accuracy with speed through a combination of a probabilistic EM and an adaptable (steerable) FBsPCA representation. This combination provides additional robustness against imperfect alignment in either translation or rotation.

An Algorithm for Single Particle Reconstruction (ASPIRE) first developed FBsPCA \cite{zhao2013fbspca} and its fast algorithm \cite{zhao2016fast} as a steerable image representation and initially a denoising technique. ASPIRE also proposed a particle averaging method \cite{zhao2014aspirerot} which computes bispectrums to find approximated nearest neighbors for each image \cite{Jones15679}. These initial nearest neighbors form a graph whose manifold is extracted via vector diffusion maps (VDM) \cite{singer2012vdm}. Each image is then aligned to its neighbors to create one average. There are also other relevant methods using FBsPCA. \cite{Ma_2020}'s one-pass algorithm similarly computes the invariant features but recovers bispectrums of the class averages by solving a non-convex optimization problem. \cite{Fan2019MVDM} proposed an extension to VDM by simultaneously building VDM graph using multiple rotation frequencies. Instead of relying on invariant features based on FBsPCA, our method exploits the advantage of FBsPCA steerability and the low-dimensionality of a learned latent EM space to quickly model the a posteriori particle projection image mixture distribution, exploiting the full FBsPCA representation.

Other prior 2D classification/clustering algorithms include those that infer the structures of projection from some prior 3D template structure. KerdenSOM \cite{PASCUALMONTANO2001233} defines a mixture distribution and learns the kernel probability neighborhood of class representations. \cite{2dcluster} assigns each projection a 3D orientation angle relative to three referential projections via a common line search of 2D Fourier transforms, accommodated by the projection-slice theorem \cite{Bracewell1956StripII}.
\section{Probabilistic PolarGMM}



3D (translation and rotation) alignment and clustering of  particle images $\{ I^{(i)} \}_{i=1}^n$ of size $L \times L$ pixels into projection clusters and aligns them with planar rotation and planar translation. In other words, let $I'(\bfx) = I(R_{\alpha} \bfx + \bft)$ be a rigidly transformed image of $I(\bfx)$, then $I$ and $I(R_{\alpha} \bfx + \bft)$ should be assigned into the same cluster. If no alignment is applied to $I$, alignment should then translate $I'$ with $-\bft$ and rotate it with $R_{-\alpha}$. Note that a correct alignment is relative between images; there is no preferred alignment for each cluster. If we write the 3D orientation in terms of pitch-yaw-roll rotation, 2D classification groups all particle images with similar pitch and yaw together, while alignment determines relative roll and planar translation. In practice, it is acceptable to predicts the rotation and translation up to some precision, i.e. giving a range of rotation and translation.

\begin{figure}
	\centering
	\includegraphics[width=0.5\linewidth,keepaspectratio]{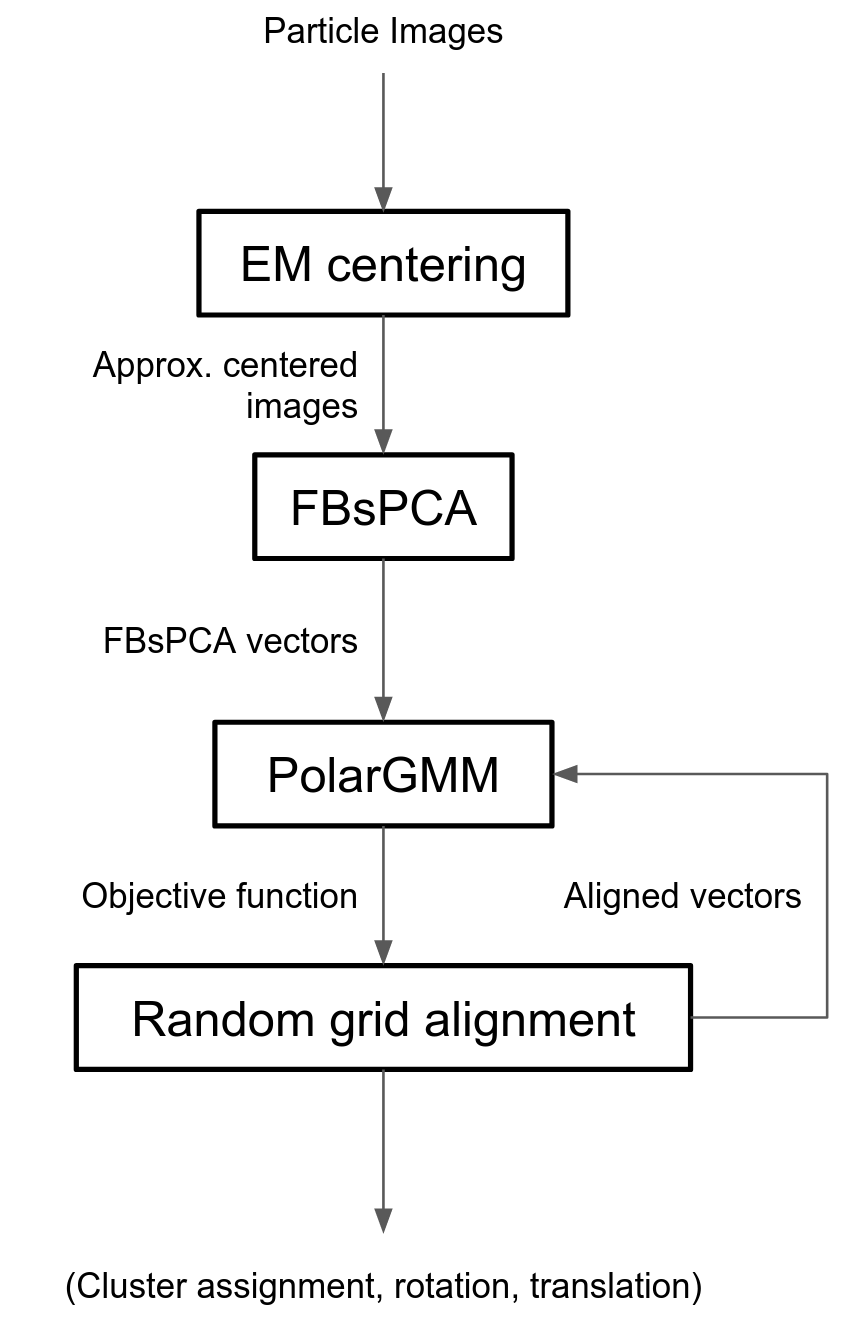}
	\caption{Probabilistic PolarGMM pipeline}
	\label{fig:pipeline}
\end{figure}

At the core, FBsPCA poses a suitable representation for solving 2D classification and alignment problem. Beyond performing computation in a compact representation of $m \ll L^2$ dimensions, its steerability further accelerates rotation, and so, speeds up the alignment search in magnitudes. Section \ref{method:fbspca} briefly introduces FBsPCA which is extended with translation operators in Section \ref{method:translation} to handle translation alignment. However, the translation operator is expensive which would increase computation cost if more translations is needed. As a remedy, Section \ref{method:emcenter} introduces a pre-centering algorithm to reduce the translation search space. Finally, Section \ref{method:polargmm} states the primary 2D classification algorithm which takes clustering and alignment imperfection into consideration. Figure \ref{fig:pipeline} illustrates the pipeline through these steps.

\subsection{Fourier-Bessel Steerable Principal Component Analysis} \label{method:fbspca}

FBsPCA is a suitable representation to solve 2D classification as it endows a computationally efficient and mathematically tractable rotation operator. We relies on this steerability mainly to simplify alignment algorithm and to incorporate its uncertainty into the mixture probability density function (pdf). Additionally, thanks to its final PCA step, FBsPCA automatically selects important features which likely correlate with the particle projection signal. After fitting FBsPCA bases $\Psi \in \mathbb{C}^{L^2 \times m}$ and the mean of images $\mu_I \in \mathbb{C}^{L^2}$, we obtain FBsPCA vectors $\{ \bfz^{(i)} \}_{i=1}^n$ with $\bfz^{(i)} \in \mathbb{C}^{m}$ from the particle image dataset $\{ \bfI^{(i)} \}_{i=1}^n$ where $\bfI^{(i)} \in \mathbb{C}^{L^2}$ using Equation \ref{eq:FBsPCA-transform-compact}. Verbally, a basis in $\Psi$ supports images in Fourier space and consists of a linear combination of Bessel functions with different frequencies along radial and angular axes. We refer to the main FBsPCA paper \cite{zhao2013fbspca} for full description of the bases and the subsequent one \cite{zhao2016fast} for the fast fitting algorithm.
\begin{equation} \label{eq:FBsPCA-transform-compact}
\bfz^{(i)} = \braket{\Psi, \bfI^{(i)} - \mu_I}
\end{equation}

Since $\Psi$ is unitary, the decoding equation from FBsPCA vector to an image follows Equation \ref{eq:FBsPCA-transform-compact-inv}.
\begin{equation} \label{eq:FBsPCA-transform-compact-inv}
\bfI^{(i)} = \braket{\Psi^{\dagger}, \bfz^{(i)}} + \mu_I
\end{equation}

We also write each FBsPCA vector $\bfz$ in polar coordinate $(\bfr, \bfphi)$ such that $j$-th coordinate is $z_j = r_j e^{\iota \phi_j}$. Additionally, the FBsPCA algorithm also gives an angular frequency vector $\bfomega$ per dataset which conveys the effect of a rotation in image space by $\alpha$ radian to the rotation in polar coordinate by Equation \ref{eq:FBsPCA-rotation-compact}. In other words, we can compute the new FBsPCA vector for a rotated image by adding the angular vector with $\alpha \bfomega$.
\begin{equation} \label{eq:FBsPCA-rotation-compact}
R_\alpha(\bfr, \bfphi) = (\bfr, \bfphi + \alpha \bfomega)
\end{equation}

\subsection{FBsPCA Translation} \label{method:translation}

There are two possible translation operators with different computation complexities, namely vanilla and cached operators. The vanilla translation operator follows the decode-translate-encode pattern (Equation \ref{eq:txy-naive}) which takes $O(mL^2)$ per operation.
\begin{equation} \label{eq:txy-naive}
T_\bft(\bfz) = \braket{\Psi, T_\bft(\braket{\Psi^{\dagger}, \bfz} + \mu_I) - \mu_I}
\end{equation}

Since rigid transformation preserves inner product, we can move $T_{\bft}$ by applying $T_{-\bft}$ on both sides. The cached translation operator utilizes this property to write an affine transformation (Equation \ref{eq:txy-dict}) and caches necessary constants. Computing $\Psi_{\bft}$ and $\mu_{\bft}$ takes $O(m^2 L^2)$ time and $O(m^2)$ space while applying translation operator takes $O(m^2)$ per operation.
\begin{equation} \label{eq:txy-dict}
\begin{split}
T_\bft(\bfz)
    &= \braket{\Psi^{\dagger} T_{-\bft}(\Psi), \bfz} + \braket{T_{-\bft}(\Psi) - \Psi, \mu_I} \\
    &= \braket{\Psi_{\bft}, \bfz} + \mu_{\bft}
\end{split}
\end{equation}

Therefore, if a translation vector is applied $\omega(m)$ times, it is worth using the cached translation operator over the vanilla one, and vice versa. In our pipeline, alignment heavily uses the cached translation operator while PolarGMM update step uses the vanilla operator.

\subsection{EM Centering} \label{method:emcenter}

EM centering coarsely aligns particle images under translation. It significantly reduces the alignment search space, accelerating mixture model fitting.

Two primary observations of a noisy particle image are; (1) pixels under the projection of the particle is darker than the background, and (2) the shape of particle projection is concentrated around its center. Therefore, to find a center, EM centering assumes that the projection creates normal-distributed pixels around a position on the image with some darker pixel value. The central position and pixel value are included in $\bfmu_s$. To incorporate noise and non-projection pixels, EM centering models every pixel in an image as if it exclusively comes from either signal distribution $f_s$ or background distribution $f_b$ in Equation \ref{eq:emcenter-pdf} where the normalization factor of the signal density depending on $\Sigma_s$ is hidden for brevity. The EM steps can be derived by optimizing the parameters $\bftheta = (\bfmu_s, \Sigma_s, \mu_b, \sigma_b)$ over the corresponding Q-function. Here we generate sample points $\bfp = (x, y, I(x, y))$ as the tuple of row, column, and pixel value of the image $I$ where $p_c = I(x, y)$. Therefore, the center of the particle can be recovered from $\bfmu_s$.
\begin{equation} \label{eq:emcenter-pdf}
\begin{split}
f_s(\bfp; \bfmu_s, \Sigma_s) &\propto \exp \left( -\frac{1}{2} (\bfp - \bfmu_s)^\T \Sigma_s^{-1} (\bfp - \bfmu_s) \right) \\
f_b(\bfp; \mu_b, \sigma_b) &= \frac{1}{\sqrt{2 \pi \sigma_b^2}} \exp \left( -\frac{1}{2} \frac{(p_c - \mu_b)^2}{\sigma_b^2} \right) \\
f(\bfp; \bftheta) &= f_s(\bfz; \bfmu_s, \Sigma_s) + f_b(\bfz; \mu_b, \sigma_b)
\end{split}
\end{equation}

\subsection{Polar-coordinated Gaussian Mixture Model} \label{method:polargmm}

Polar-coordinated Gaussian mixture model (PolarGMM) implicitly assigns each sample with predicted alignment angle $\tilde{\alpha}$ into all rotation bins with normal distribution centered around $\tilde{\alpha}$. This choice of assignment models the imperfection in the alignment with some global uncertainty. However, it is equivalent and more straightforward to convolve the assignment distribution into individual cluster pdf. Equations \ref{eq:polargmm-pdf} present an approximation of such pdf $f_c$ and the mixture pdf $f$ with parameters $\bftheta = (\bfpi, M^{(1)}, \dots, M^{(C)})$ and $M^{(c)} = (\bfmu_r^{(c)}, \bfmu_\phi^{(c)}, \bfsigma_r^{(c)}, \bfsigma_\phi^{(c)})$ where the normalization factor depending on all $\sigma_{r, j}$ and $\sigma_{\phi, j}$ is hidden once again.
\begin{equation} \label{eq:polargmm-pdf}
\begin{split}
f_c(\bfr, \bfphi; M) &\propto
    \exp \left( -\sum_{j=1}^m \frac{(r_j - \mu_{r,j})^2}{ 2 \sigma_{r,j}^2} + \frac{\left| \phi_j - \mu_{\phi,j}\right|^2}{ 2 \sigma_{\phi,j}^2} \right) \\
f(\bfr, \bfphi; \bftheta) &= \sum_{c=1}^C \pi^{(c)} f_c(\bfr, \bfphi; M^{(c)})
\end{split}
\end{equation}

Explicitly, a system of formulae in Equation \ref{eq:polargmm-update} shows the EM steps derived from the mixture pdf, given aligned samples $(\tilde{\bfr}^{(i)}, \tilde{\bfphi}^{(i)}) = R_{\tilde{\alpha}^{(i)}} T_{\tilde{\bft}^{(i)}} (\bfr^{(i)}, \bfphi^{(i)})$. 

\begin{equation} \label{eq:polargmm-update}
\begin{split}
\Pr(c | \bfr, \bfphi, \bftheta) &= \frac{\pi^{(c)} f_c( \bfr, \bfphi; M^{(c)})}{\sum_{c'=1}^C \pi^{(c')} f_{c'}( \bfr, \bfphi; M^{(c')})} \\
w_{i, c} &= \Pr(c | \tilde{\bfr}^{(i)}, \tilde{\bfphi}^{(i)}, \bftheta_0) \\
\pi^{(c)} &= \frac{1}{n} \sum_{i=1}^n w_{i, c} \\
\mu_{r,j}^{(c)} &= \frac{\sum_{i=1}^n w_{i, c} \tilde{r}_{j}^{(i)}}{\sum_{i=1}^n w_{i, c}} \\
\mu_{\phi,j}^{(c)} &= \arctan_2 \left( \frac{\sum_{i=1}^n w_{i, c} \sin(\tilde{\phi}_{j}^{(i)})}{\sum_{i=1}^n w_{i, c}}, \right. \\
& \hspace{5.2em} \left. \frac{\sum_{i=1}^n w_{i, c} \cos(\tilde{\phi}_{j}^{(i)})}{\sum_{i=1}^n w_{i, c}} \right) \\
\sigma_{r,j}^{(c)} &= \sqrt{\frac{\sum_{i=1}^n w_{i, c} (\tilde{r}_{j}^{(i)} - \mu_{r,j}^{(c)})^2}{\sum_{i=1}^n w_{i, c}}} \\
\sigma_{\phi,j}^{(c)} &= \sqrt{\frac{\sum_{i=1}^n w_{i, c} |\tilde{\phi}_{j}^{(i)} - \mu_{\phi,j}^{(c)}|^2}{\sum_{i=1}^n w_{i, c}}}
\end{split}
\end{equation}

To speed up the learning, we uniformly randomly select $B$ out of $n$ samples as a sample batch to align and fit PolarGMM in each iteration. Note that Equation \ref{eq:polargmm-update} does not take the sample batch into account for brevity.

\subsection{Random Grid Alignment} \label{method:grid-align}

To align each sample, we search for the best rotation angle and translation vector with respect to the mixture pdf over an equally spaced grid. We uniformly randomly rotate the FBsPCA vectors before performing the grid search to enlarge the set of possible rotation angles. Therefore, a rotation angle sample is $\tilde{\alpha}^{(i)} \in \{ \frac{2 \pi k}{n_\alpha} + \alpha_0 \; | \; k \in [n_\alpha], \alpha_0 \sim \text{Unif}[0, 2 \pi] \}$ where $n_\alpha$ is the grid size. Due to its smaller discrepancy between its samples, such grid search empirically performs better than uniform sampling.

Similarly, we sample translation from a circular grid with a limited radius. The grid consists of multiple rings of particular radii where each ring consists of multiple samples equally distributed in distance across all rings: $\tilde{\bft}^{(i)} \in S(R, n_r) = \{ (r \cos \theta, r \sin \theta) \; | \; r = \frac{R i}{n_r}, i \in [n_r], \theta = \frac{2 \pi j}{n_{i, \theta}}, j \in [n_{i, \theta}], n_{i, \theta} = \lfloor \frac{\pi}{\arcsin(0.5 / i)} \rfloor \}$ where $R$ is the radius limitation and $n_r$ is the number of rings in this grid. This translation grid exploits the cached translation operator in Section \ref{method:translation} by precomputing instantiated operators at each translation position in the grid.

Collectively, Algorithm \ref{algo:polargmm} outlines the proposed PolarGMM algorithm for 2D classification.

\begin{algorithm}
	\caption{PolarGMM 2D Classification on FBsPCA}
	\label{algo:polargmm}
	\begin{algorithmic}[1]
    	\State Center images $\bfI^{(i)}$ by EM centering
    	\State Fit \cite{zhao2013fbspca} and transform images into FBsPCA vectors $(\bfr^{(i)}, \bfphi^{(i)})$
		\For {$\text{iteration } =1, \dots, n_{\text{iter}}$}
			\State $\tilde{\alpha}^{(i)}, \tilde{\bft}^{(i)} = \argmax_{\alpha, \bft} f(R_\alpha T_\bft(\bfr^{(i)}, \bfphi^{(i)}); \bftheta)$
			\State Assign $w_{i, c}$ and update $\bfmu_r^{(c)}$, $\bfmu_\phi^{(c)}$, $\bfsigma_r^{(c)}$, $\bfsigma_\phi^{(c)}$, and $\bfpi^{(c)}$
		\EndFor
		\State Use $\tilde{\alpha}^{(i)}, \tilde{\bft}^{(i)}$ to align and compute averages
	\end{algorithmic} 
\end{algorithm}

\subsection{Complexity} \label{method:complexity}

In evaluation of time complexity, recall $n$ the number of images, $L$ width and height of a particle image, $m$ the number of FBsPCA components, $C$ the number of clusters, $B$ the batch size per iteration, $n_\alpha$ the rotation sample size, $n_t = |S(R, n_r)|$ the translation sample size, $n_{\text{citer}}$ the number of iterations in EM centering, and $n_{\text{iter}}$ the number of iterations of PolarGMM EM step and alignment. Table \ref{tab:complexity} summarizes time complexity in all steps. When no translation is present, this would be simplified, that is, $O(n_{\text{citer}} n L^2 + n L^3 + L^4 + n_{\text{iter}} n_\alpha m B C + n_\alpha m n C)$.
\begin{table}
    \begin{center}
    \begin{tabular}{lll}
        \toprule
        Step                                    & Time Complexity &  \\
        \midrule
        EM Centering                            & $O(n_{\text{citer}} n L^2)$ & \\
        FBsPCA                                  & $O(n L^3 + L^4)$ & \\
        Initialize $\Psi_\bft$ and $\bfmu_\bft$    & $O(n_t m^2 L^2)$ & \\
        PolarGMM                     & $O(m L^2 B + m B C)$ & \multirow{3}{*}{\hspace{-1em}$\left.\begin{array}{l}
        \\
        \\
        \\
        \end{array}\right\rbrace \times n_{\text{iter}}$} \\
        Compute $T_\bft(\bfmu)$                 & $O(n_t m^2 C)$ & \\
        Grid alignment                   & $O(n_t n_\alpha m B C)$ & \\
        Prediction                              & $O(n_t n_\alpha m n C)$ & \\
        \bottomrule
    \end{tabular}
    \end{center}
    
    \caption{Time Complexity for Each Step in PolarGMM}
    \label{tab:complexity}
\end{table}
\section{Experiments}

\begin{figure*}[t]
	\centering
	\begin{subfigure}[b]{0.24\linewidth}
	    \centering
	    \includegraphics[height=3cm,angle=0,keepaspectratio]{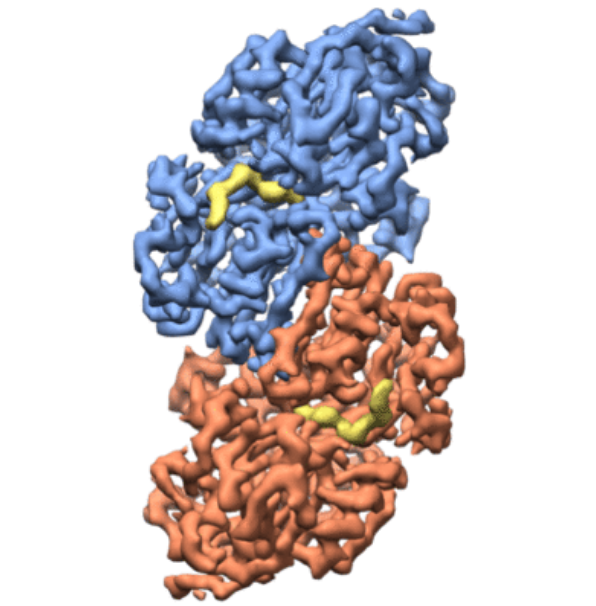}
	    \caption{70S}
	\end{subfigure}
	\begin{subfigure}[b]{0.24\linewidth}
	    \centering
	    \includegraphics[height=3cm,keepaspectratio]{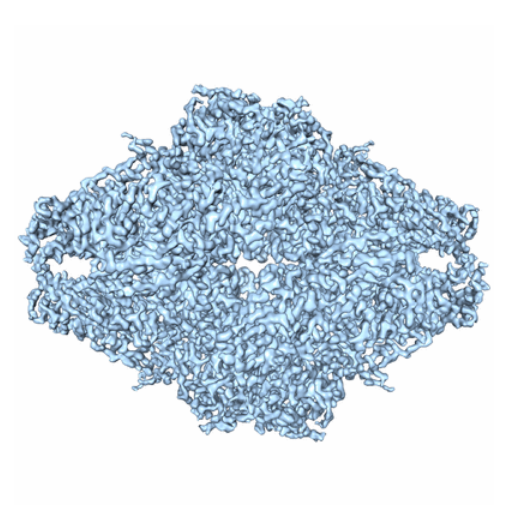}
	    \caption{Bgal}
	\end{subfigure}
	\begin{subfigure}[b]{0.24\linewidth}
	    \centering
	    \includegraphics[height=3cm,angle=0,keepaspectratio]{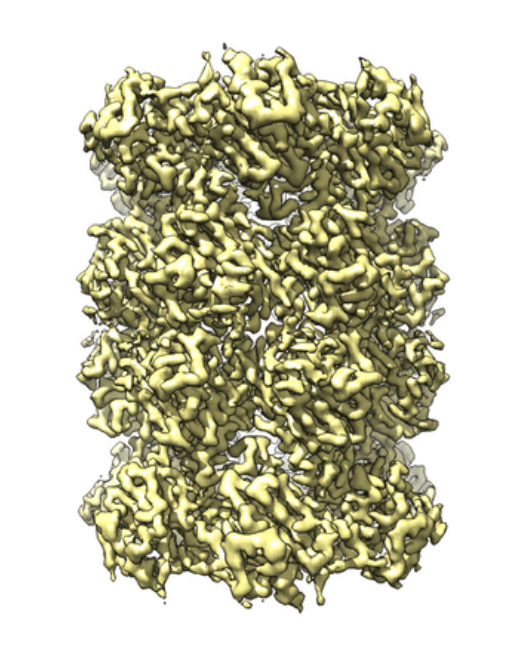}
	    \caption{T20}
	\end{subfigure}
	\begin{subfigure}[b]{0.24\linewidth}
	    \centering
	    \includegraphics[height=3cm,keepaspectratio]{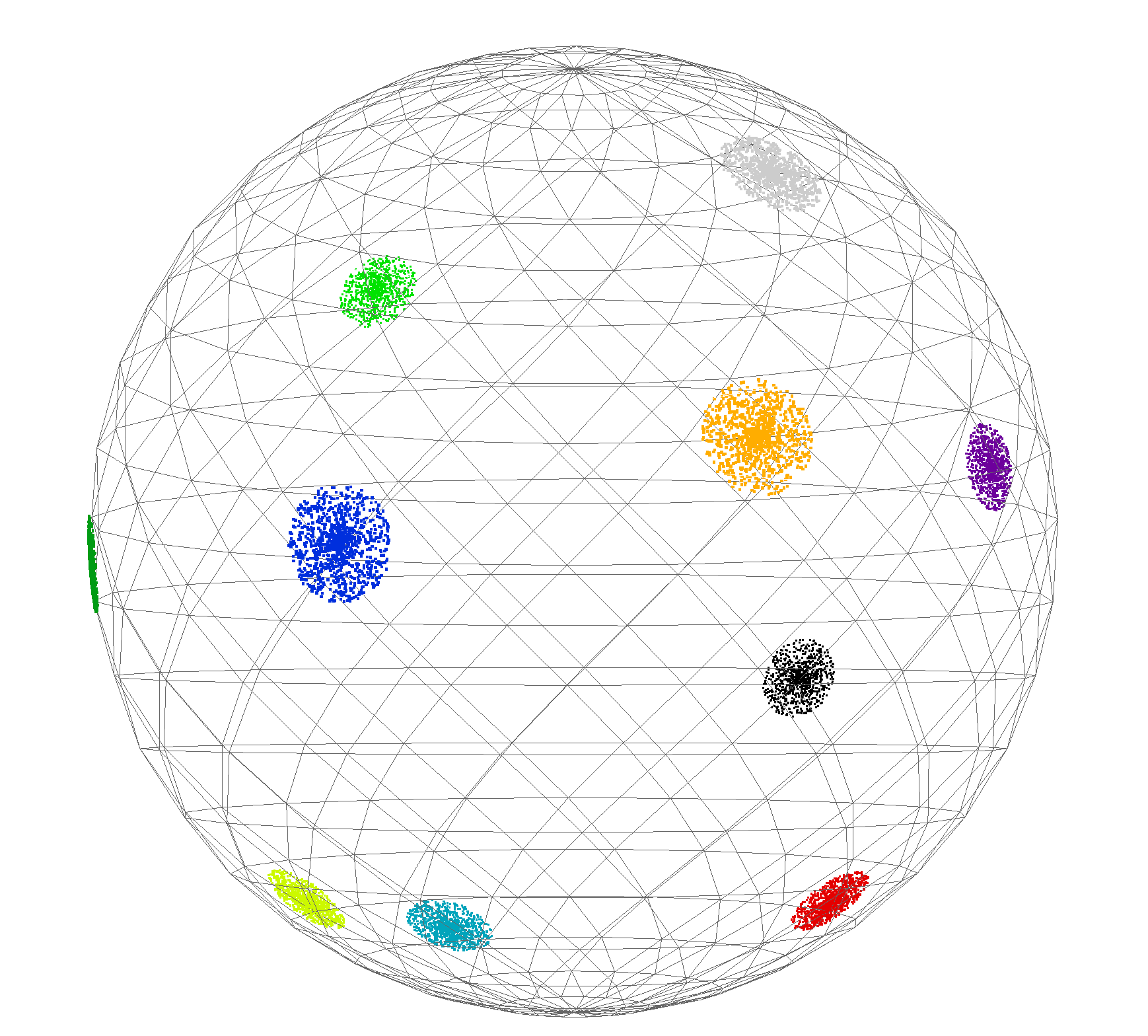}
	    \caption{Projection Directions}
	\end{subfigure}
    \caption{3D models of different particle models from their respective EMDR webpages. Colors in (d) indicate true cluster labels.}
	\label{fig:models}
\end{figure*}

We empirically measure the performance of PolarGMM in comparison with EMAN2 \texttt{e2refine2d.py} and the latest RELION 3.1 \texttt{relion\_refine\_mpi}. In order to compute clustering and alignment error metrics, we generated datasets from projection simulation to acquire true hidden labels and orientations, we couldn't otherwise from raw Cryo-EM micrographs.


\subsection{Dataset Simulation} \label{result:dataset}

Three particle models are used for testing: EMBL-EBI: 2.9-{\AA} horse liver alcohol dehydrogenase (EMD-0406, 70S) \cite{Herzik2019_70s}, 3.3-{\AA} archaeal 20S proteasome (EMD-5623), T20) \cite{Li2013_t20}, and 3.2-{\AA} beta-galactosidase (EMD-5995, Bgal) \cite{Bartesaghi2014bgal}. Figure \ref{fig:models} provides a sketch of all three particle models.

For each particle, we generate two datasets whose only difference is that the second contains random translation while the first doesn't. Per each dataset, we sample 10 cluster orientations where we apply jittering around each of the cluster orientation to create $10^4$ orientations in total (i.e. $10^3$ samples for cluster). To generate a particle image from each orientation, we project 3D volume model onto 2D plane orthographically along the orientation direction via \texttt{vtk8.2.0} ray casting. The image size is $150 \times 150$ pixels with an appropriate scaling factor such that the projections of the particles lie in some circle fitted in the frame. Finally, we apply uniformly random planar rotation over $[-\pi, \pi)$, uniformly random planar translation over disk $\{(x, y) \; | \; x^2 + y^2 \leq 15^2\}$ (second dataset only), and i.i.d. pixel Gaussian noises with SNR $1/5$.

For brevity, we denote 70S, Bgal, and T20 as the corresponding datasets without random planar translation. Likewise, 70S-T, Bgal-T, and T20-T refer to the datasets with random planar translation.


\subsection{Metrics} \label{result:metrics}

With known true cluster labels and orientations of particle projections, our experiments measure clustering performance as well as alignment errors. The cluster metrics are imported from scikit-learn library \cite{scikit-learn}.

\paragraph{Accuracy (ACC)} measures the best cluster indices matching between clusterings $C$ (true label) and $K$ (predicted label).
\begin{equation} \label{eq:metrics-acc}
ACC = \max_m \frac{\sum_{i=1}^n \mathbbm{1}_{\{C_i = m(K_i)\}}}{n}
\end{equation}
	
\paragraph{Adjusted Mutual Information (AMI)} \cite{StrehlG02mi} considers cluster assignments (true and predicted assignments) as two categorical distributions and measures their mutual information. AMI is adjusted to measure zero from uniformly random assignment where we denotes $E[MI]$ as the . Let $H(X)$ be the information entropy of random variable $X$ and $MI(X, Y)$ be the mutual information between random variables $X$ and $Y$.
\begin{equation} \label{eq:metrics-mi}
\begin{aligned}
AMI = \frac{MI(C, K) - E[MI]}{\text{mean}(H(C), H(K)) - E[MI]}
\end{aligned}
\end{equation}

\paragraph{Homogeneity (h) and Completeness (c)} \cite{rosenberg2007vMeasure} focus on the conditional entropy of one clustering given another clustering. A high homogeneity indicates that each predicted cluster contains only samples from the same true cluster. On the other hand, a high completeness indicates that samples in the same true cluster are assigned to the same predicted cluster.
\begin{equation} \label{eq:metrics-hcv}
\begin{aligned}
h = 1 - \frac{H(C | K)}{H(K)}, \qquad
c = 1 - \frac{H(K | C)}{H(C)}
\end{aligned}
\end{equation}
	
\paragraph{Relative Angle Squared Error (AE-2) and Relative Translation Squared Error (TE-2)} measures the relative alignment errors on pairs of samples that are classified similarly. The prediction's error is measured based on the known true relative planar rotation. We measure AE-2 in radian and TE-2 in pixel unit length. Let $N = N(C, K)$ be the number of jointly assigned sample pairs between true and predict clusterings, $N = \sum_{i \neq j} \mathbbm{1}_{i, j}$ where $\mathbbm{1}_{i, j} = \mathbbm{1}_{\{C_i = C_j \land K_i = K_j\}}$.
\begin{equation} \label{eq:metrics-relerr}
\begin{aligned}
\text{AE-2} = \left( \frac{1}{N} \sum_{i \neq j} \mathbbm{1}_{i, j} | (\alpha^{(*)}_i - \alpha^{(*)}_j) - (\tilde{\alpha}_i - \tilde{\alpha}_j) |^2  \right)^{1/2} \\
\text{TE-2} = \left( \frac{1}{N} \sum_{i \neq j} \mathbbm{1}_{i, j} \norm{ (\bft^{(*)}_i - \bft^{(*)}_j) - (\tilde{\bft}_i - \tilde{\bft}_j) }^2_2 \right)^{1/2}
\end{aligned}
\end{equation}


\subsection{Configurations} \label{result:config}

We set FBsPCA particle radius ratio and truncation parameters to $0.6$ and $10.0$ respectively for 70S, Bgal, and T20. For datasets with translation 70S-T, Bgal-T, and T20-T, we set them to $0.8$ and $10.0$. We select only $m = 50$ top FBsPCA components to represent each image. As a reminder, there are $L^2 = 150^2$ pixels in an image from these datasets. Other parameter settings are $C = 10$, $B = 5000$, $n_t = 60$, $n_\alpha = 60$, $n_{\text{citer}} = 10$, and $n_{\text{iter}} = 10$.

The experiment environment has Intel Xeon CPU E5-1660 with 16 CPUs at 3.20 Hz. Although the system has 64 GB of RAM, the memory usage in these experiments never allocate more than 5 GB.


\subsection{2D Classification Results} \label{result:result}

We summarize the quantitative results with expectation statistics across experiment runs where standard deviations are adjusted accordingly to the number of experiment trials. We also highlight scores that have the best mean in each column, that is, highest clustering metrics and lowest alignment errors. Tables \ref{table:result-no-t} and \ref{table:result-t} show the results when random translation is not presented and is presented respectively.

PolarGMM consistently outperforms EMAN2 and RELION in terms of accuracy, AMI, and homogeneity, with only one exception (AMI in T20-T). It recovers the clustering up to $86\%$ accuracy, $94\%$ AMI, and $91\%$ homogeneity on 70S dataset with the worst accuracy at $51\%$ accuracy, $64\%$ AMI, and $63\%$ homogeneity in T20-T due to the particle's higher symmetry. (We dont assume we know the particle's symmetries, unlike EMAN2 and RELION).

PolarGMM achieves higher completeness scores as well. When RELION reports better completeness measures, it is often due to the collapse in its number of clusters. This collapse of course leads to lower resolution in the 3D reconstruction. In other words, RELION can choose to predict a fewer number of projection clusters than specified and reequired for the best 3D reconstruction. Decreasing the number of projection clusters thus artificially boosts the completeness score.

In terms of translation alignment errors, EMAN2 performs slightly better than PolarGMM on 70S-T and Bgal-T, albeit with small margins on pixel counts. Recall that rotation and translation errors are measured in degrees and pixels (respectively). For datasets with synthetically generated translations (shifts within each boxed image), PolarGMM's alignments comprise a maximum of around 1 degree of rotation and  2-5 pixels translation squared deviations.

\begin{table*}[bt]
\begin{center}
\begin{tabular}{l cc cc c}
    \toprule
    \multirow{2}{*}{Methods} &
    \multicolumn{5}{c}{70S} \\
		& ACC & AMI & h & c & AE-2 \\
		\midrule
		PolarGMM &
			$\bf 0.86 \pm 0.02$ & $\bf 0.94 \pm 0.01$ & $\bf 0.91 \pm 0.01$ & $\bf 0.97 \pm 0.00$ & $\bf 0.05 \pm 0.00$ \\
		EMAN2 &
			$0.72 \pm 0.02$ & $0.84 \pm 0.02$ & $0.82 \pm 0.02$ & $0.86 \pm 0.02$ & $1.71 \pm 0.05$ \\
		RELION &
			$0.60 \pm 0.03$ & $0.81 \pm 0.02$ & $0.72 \pm 0.02$ & $0.94 \pm 0.01$ & $0.20 \pm 0.05$ \\
    \bottomrule
\end{tabular}

\begin{tabular}{l cc cc c}
    \toprule
    \multirow{2}{*}{Methods} &
    \multicolumn{5}{c}{Bgal} \\
		& ACC & AMI & h & c & AE-2 \\
		\midrule
		PolarGMM &
			$\bf 0.82 \pm 0.02$ & $\bf 0.90 \pm 0.01$ & $\bf 0.86 \pm 0.01$ & $\bf 0.94 \pm 0.00$ & $\bf 0.07 \pm 0.01$ \\
		EMAN2 &
			$0.41 \pm 0.02$ & $0.53 \pm 0.02$ & $0.52 \pm 0.02$ & $0.54 \pm 0.02$ & $0.78 \pm 0.08$ \\
		RELION &
			$0.59 \pm 0.05$ & $0.77 \pm 0.02$ & $0.73 \pm 0.03$ & $0.83 \pm 0.01$ & $0.09 \pm 0.00$ \\
    \bottomrule
\end{tabular}

\begin{tabular}{l cc cc c}
    \toprule
    \multirow{2}{*}{Methods} &
    \multicolumn{5}{c}{T20} \\
		& ACC & AMI & h & c & AE-2 \\
		\midrule
		PolarGMM &
			$\bf 0.76 \pm 0.02$ & $\bf 0.87 \pm 0.01$ & $\bf 0.84 \pm 0.01$ & $0.91 \pm 0.01$ & $\bf 0.68 \pm 0.02$ \\
		EMAN2 &
			$0.48 \pm 0.01$ & $0.59 \pm 0.01$ & $0.58 \pm 0.01$ & $0.61 \pm 0.01$ & $1.78 \pm 0.04$ \\
		RELION &
			$0.46 \pm 0.02$ & $0.73 \pm 0.01$ & $0.59 \pm 0.02$ & $\bf 0.97 \pm 0.01$ & $0.90 \pm 0.09$ \\
    \bottomrule
\end{tabular}
\end{center}

\caption{Means and their standard deviations of metrics on datasets without random translation}
\label{table:result-no-t}
\end{table*}

\begin{table*}[bt]
\begin{center}
\begin{tabular}{l cc cc cc}
    \toprule
    \multirow{2}{*}{Methods} &
    \multicolumn{6}{c}{70S-T} \\
		& ACC & AMI & h & c & AE-2 & TE-2 \\
		\midrule
		PolarGMM &
			$\bf 0.81 \pm 0.02$ & $\bf 0.87 \pm 0.01$ & $\bf 0.86 \pm 0.01$ & $\bf 0.88 \pm 0.01$ & $\bf 0.73 \pm 0.01$ & $2.83 \pm 0.03$ \\
		EMAN2 &
			$0.76 \pm 0.02$ & $0.86 \pm 0.01$ & $0.85 \pm 0.01$ & $\bf 0.88 \pm 0.01$ & $2.06 \pm 0.03$ & $\bf 2.66 \pm 0.55$ \\
		RELION &
			$0.57 \pm 0.03$ & $0.75 \pm 0.01$ & $0.69 \pm 0.02$ & $0.82 \pm 0.01$ & $1.17 \pm 0.08$ & $8.28 \pm 0.15$ \\
    \bottomrule
\end{tabular}

\begin{tabular}{l cc cc cc}
    \toprule
    \multirow{2}{*}{Methods} &
    \multicolumn{6}{c}{Bgal-T} \\
		& ACC & AMI & h & c & AE-2 & TE-2 \\
		\midrule
		PolarGMM &
			$\bf 0.67 \pm 0.02$ & $\bf 0.73 \pm 0.01$ & $\bf 0.72 \pm 0.01$ & $\bf 0.74 \pm 0.01$ & $\bf 1.03 \pm 0.01$ & $5.43 \pm 0.04$ \\
		EMAN2 &
			$0.47 \pm 0.03$ & $0.58 \pm 0.04$ & $0.57 \pm 0.04$ & $0.59 \pm 0.04$ & $1.50 \pm 0.05$ & $\bf 5.33 \pm 0.82$ \\
		RELION &
			$0.34 \pm 0.01$ & $0.42 \pm 0.01$ & $0.41 \pm 0.01$ & $0.44 \pm 0.01$ & $1.32 \pm 0.04$ & $9.53 \pm 0.16$ \\
    \bottomrule
\end{tabular}

\begin{tabular}{l cc cc cc}
    \toprule
    \multirow{2}{*}{Methods} &
    \multicolumn{6}{c}{T20-T} \\
		& ACC & AMI & h & c & AE-2 & TE-2 \\
		\midrule
		PolarGMM &
			$\bf 0.51 \pm 0.01$ & $0.64 \pm 0.01$ & $\bf 0.63 \pm 0.01$ & $0.66 \pm 0.01$ & $\bf 1.11 \pm 0.02$ & $\bf 2.50 \pm 0.02$ \\
		EMAN2 &
			$0.44 \pm 0.02$ & $0.62 \pm 0.01$ & $0.60 \pm 0.02$ & $0.63 \pm 0.01$ & $1.67 \pm 0.06$ & $3.77 \pm 0.71$ \\
		RELION & 
		    $0.47 \pm 0.01$ & $\bf 0.69 \pm 0.01$ & $0.60 \pm 0.02$ & $\bf 0.80 \pm 0.02$ & $1.35 \pm 0.08$ & $6.45 \pm 0.30$ \\
    \bottomrule
\end{tabular}
\end{center}
\caption{Means and their standard deviations of metrics on dataset with random translation}
\label{table:result-t}
\end{table*}

\begin{figure}
	\centering
	\includegraphics[width=\linewidth,keepaspectratio]{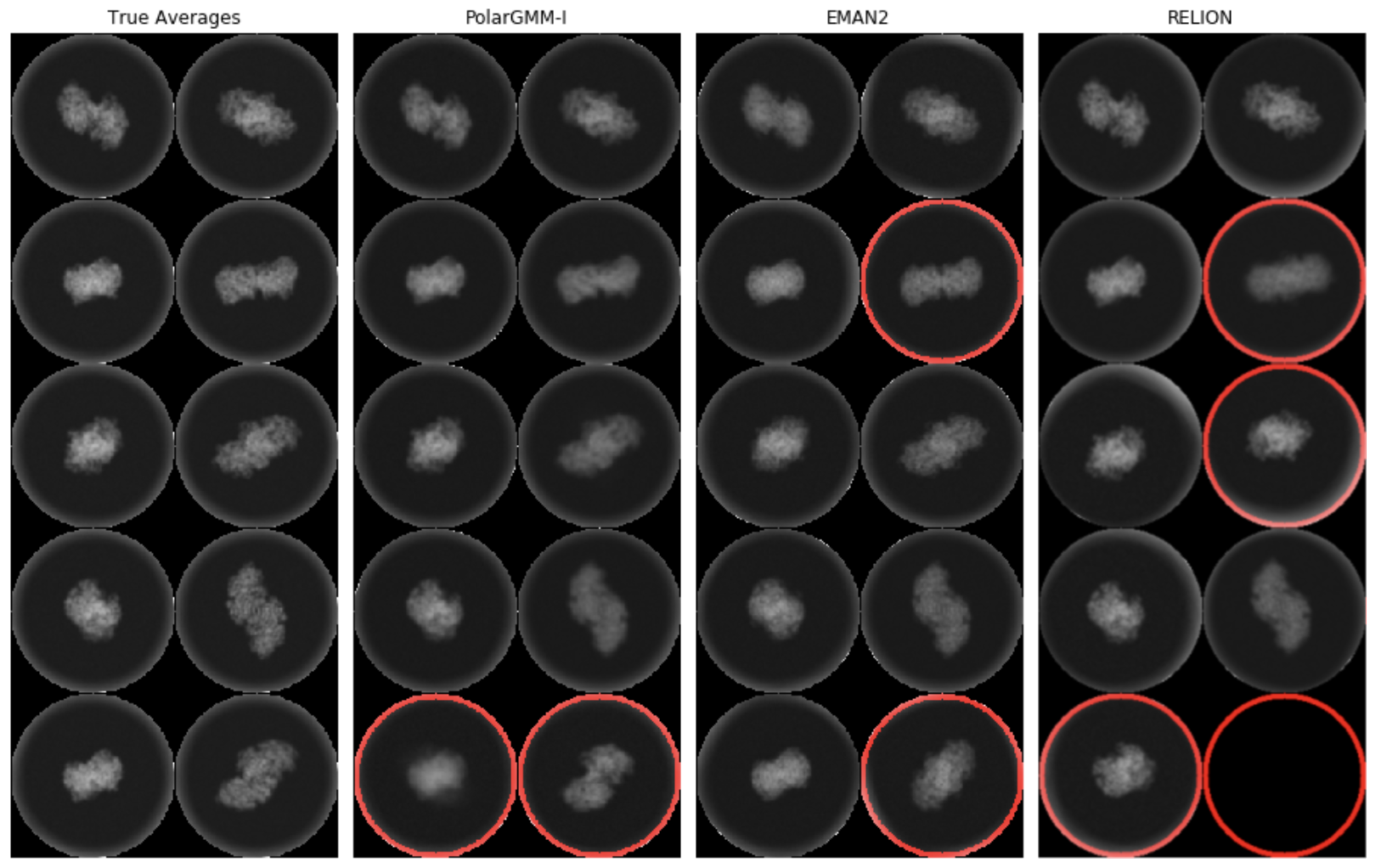}
	\caption{Class averages from 70S-T. Averages with faulty features, blurry, or missing are marked with red circles}
	\label{fig:clsavg_70s}
\end{figure}

\begin{figure}
	\centering
	\includegraphics[width=\linewidth,keepaspectratio]{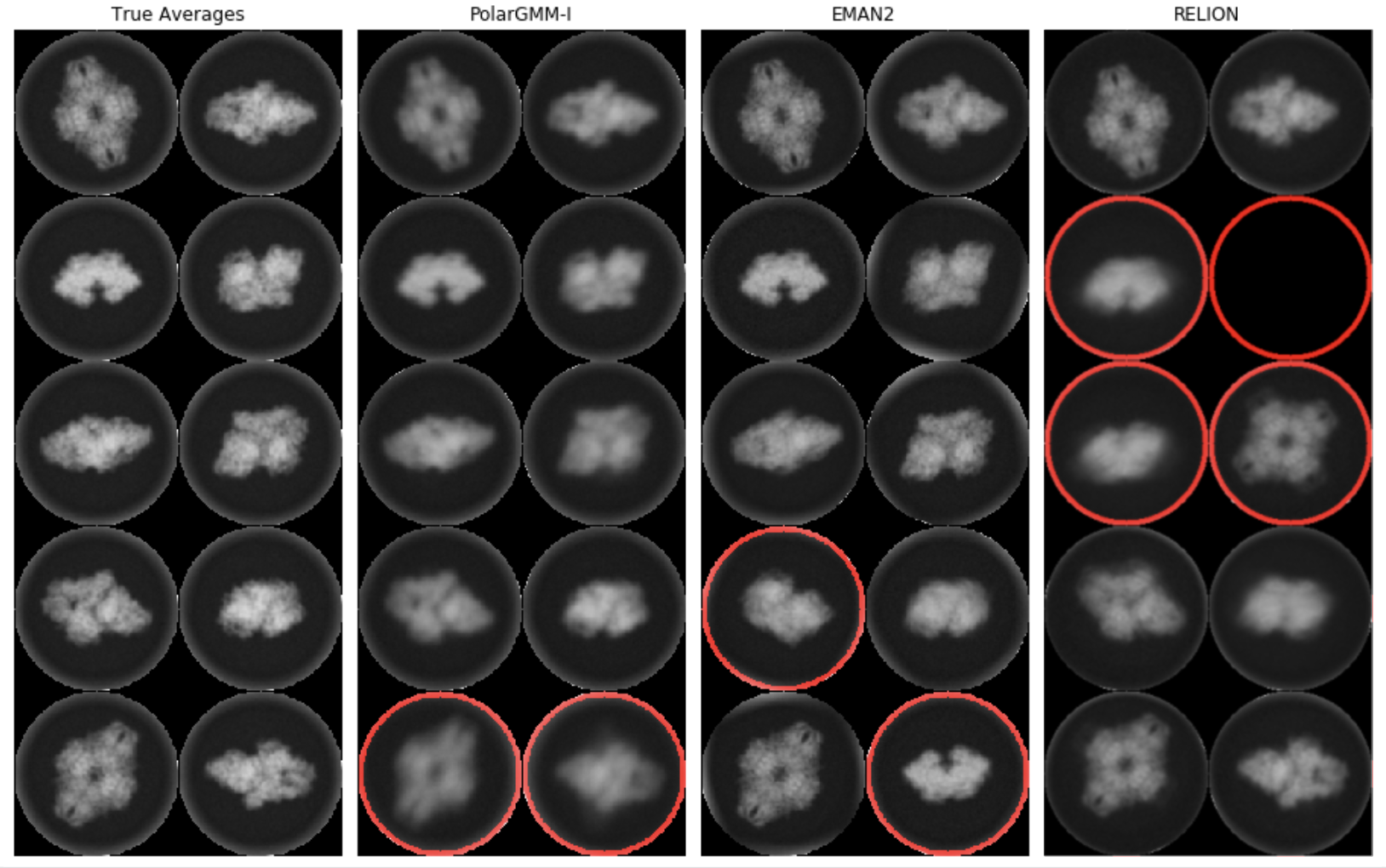}
	\caption{Class averages from Bgal-T. Averages with faulty features, blurry, or missing are marked with red circles}
	\label{fig:clsavg_beta-g}
\end{figure}

\begin{figure}
	\centering
	\includegraphics[width=\linewidth,keepaspectratio]{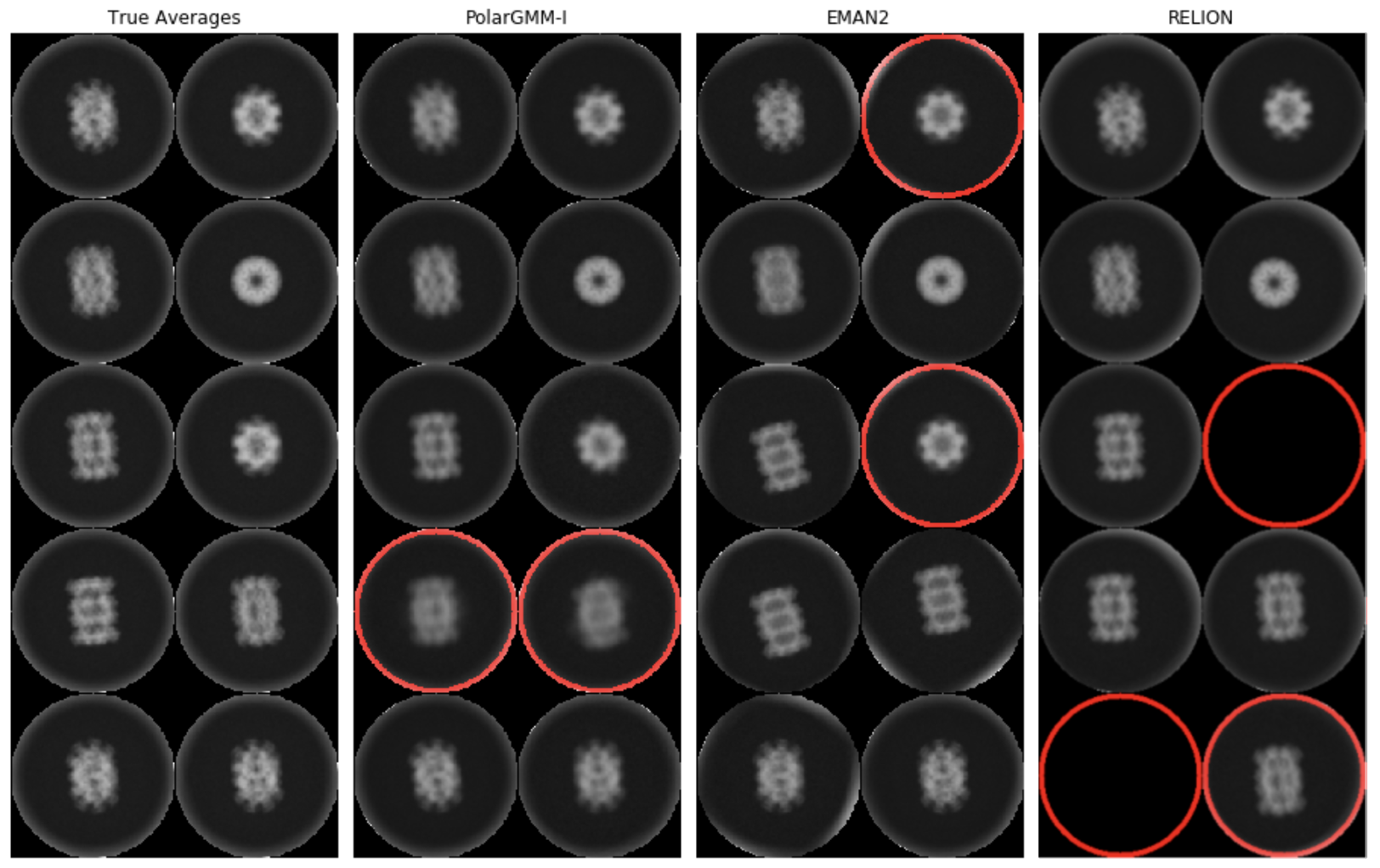}
	\caption{Class averages from T20-T. Averages with faulty features, blurry, or missing are marked with red circles}
	\label{fig:clsavg_t20}
\end{figure}

To visualize 2D classification and averaging, Figures \ref{fig:clsavg_70s}, \ref{fig:clsavg_beta-g}, and \ref{fig:clsavg_t20} show the selected class averages from one of the PolarGMM runs in comparison with the ideal averages. These are selected and rearranged to match similar averages together. For PolarGMM and EMAN2, we choose the sets of class averages which has the highest number of images distinctively matched with the true averages. For RELION, we select such sets of class averages which also has the highest number of produced averages. We highlight some averages with a noticeable defect, which is admittedly subjective.

\subsection{Computation Costs} \label{result:cost}

Table \ref{tab:exe-time} shows approximated computation times on the same system. In the case of no translation prediction, PolarGMM's total execution time is around 40 minutes while RELION and EMAN2 take 4-5 times longer. However, when a random translation is presented, the performance is around the same as RELION but still twice as fast as EMAN2. Keep in mind that our methods are implemented solely in Python whereas RELION is written in C++ while EMAN2 implements low-level operations in C++.

\begin{table}
    \centering
    \begin{tabular}{lcc}
        \toprule
        Steps/Methods                                 & 70S/Bgal/T20 & 70S-T/Bgal-T/T20-T \\
        \midrule
        EM Centering                            & - & 0.2 \\
        FBsPCA                                  & 0.5 & 0.5 \\
        PolarGMM                                & 0.2 & 3.3 \\
        \midrule
        EMAN2                                   & 3.3 & 5.8 \\
        RELION                                  & 3.0 & 3.6 \\
        \bottomrule
    \end{tabular}
    \caption{Mean execution time (in hours) across steps and datasets}
    \label{tab:exe-time}
\end{table}
\section{Conclusion}

FBsPCA's steerability provides an efficient way  to measure rotational distances between particle projections images. Coupled to unsupervised latent space soft clustering this yields a separation of  projection mixture clusters, as required for Cryo-EM 3D reconstruction. Our  PolarGMM framework thus combines the best of EM translation centering, and iterative FBsPCA rotation clustering to robustly solve the unknown pose, 3D noisy particle projection alignment problem.  The results obtained with PolarGMM are comparable in clustering accuracy  and  superior in efficiency, when compared with the implemented techniques in popular cryo-EM tools EMAN2 and RELION. 

PolarGMM, however can be improved for aligning particle projections when translation distances  are large . This however is not the case in practice as particles are carefully boxed from the single electron micrograph projection, prior to particle alignment and clustering. Future works would primarily aim to use both a product of steerable and ``shiftable'' bases (such as Wigner-D) combining translation and rotation into psuedo-polar space. Other directions of research include formal analysis of PolarGMM's tradeoffs and error bounds.

\section*{Acknowledgement}

This research was supported in part by a grant from NIH - R01GM117594, by the Peter O’Donnell Foundation and in part from a grant from the Army Research Office accomplished under Cooperative Agreement Number W911NF-19-2-0333. The views and conclusions contained in this document are those of the authors and should not be interpreted as representing the official policies, either expressed or implied, of the Army Research Office or the U.S. Government. The U.S. Government is authorized to reproduce and distribute reprints for Government purposes notwithstanding any copyright notation herein.

{\small
\bibliographystyle{ieee_fullname}
\bibliography{egbib}
}

\clearpage
\appendices

\section{FBsPCA}

Given an image $I^{(i)}$ with the content of interest lying in a circle of radius $L$ centered at the origin, we can decompose its Fourier transform $F(I^{(i)})$ with a family of $\gamma$-band-limited Fourier-Bessel functions as in Equation \ref{eq:FBsPCA-image}. The constants $k_{\max}$ and $p_k$ are assigned based on particle radius and the Fourier band-limiting parameter $\gamma$; more details on approximations of Fourier-Bessel functions  based on $k_{\max}$ are given for examples in reference\cite{zhao2013fbspca}.
\begin{equation} \label{eq:FBsPCA-image}
F(I^{(i)})(r, \phi) = \sum_{k = -k_{\max}}^{k_{\max}} \sum_{q = 1}^{p_k} a^{(i)}_{k, q} \psi^{k, q}(r, \phi)
\end{equation}
\begin{equation} \label{eq:FBsPCA-psi}
\psi^{k, q}(r, \phi) = \begin{cases}
	N_{k, q} J_k(R_{k, q} \frac{r}{\gamma}) e^{\iota k \phi}, & r \leq \gamma \\
	0, & r > \gamma
\end{cases}
\end{equation}

where $\psi^{k, q}$ is the Fourier-Bessel basis defined in Equation \ref{eq:FBsPCA-psi} and $a^{(i)}_{k, q} \in \mathbb{C}$ is the corresponding coefficient. Here $N_{k, q}$ is the normalization term such that $\int_{0}^{2 \pi} \int_{0}^\gamma |\psi^{k, q} |^2 dr d\theta = 1$. $J_k$ is the Bessel function of the first kind with integer parameter $k$. Finally, $R_{k, q}$ is the $q$-th root such that $J_k(R_{k, q}) = 0$.

\begin{figure}[hbt]
	\centering
	\includegraphics[width=6cm,height=6cm,keepaspectratio]{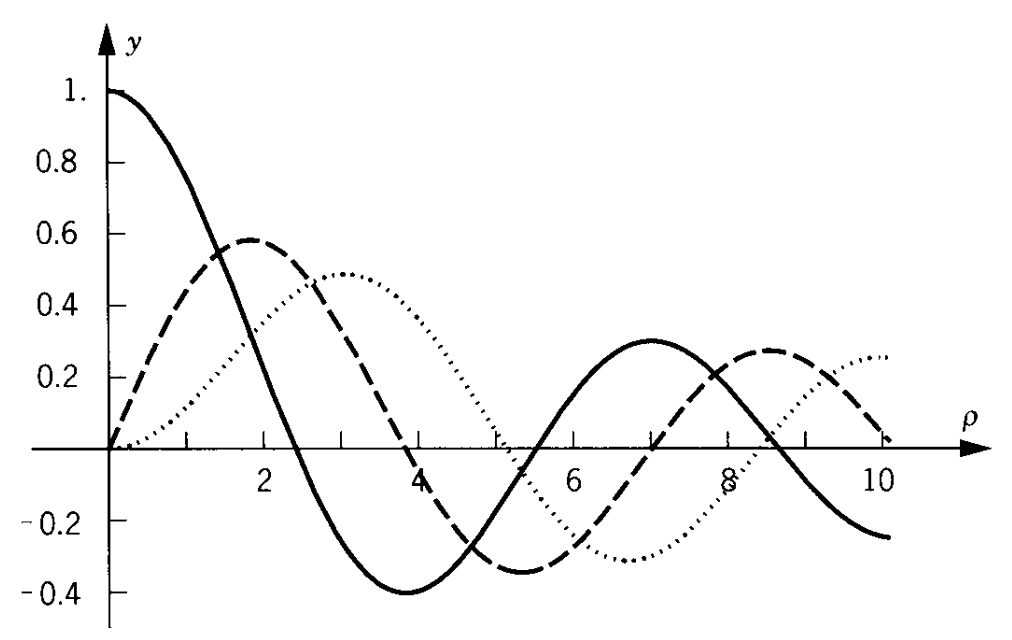}
	\caption{Bessel function of the first kind from \cite{strauss2008pde}. Solid, dashed, and dotted lines show $J_0$, $J_1$, and $J_2$ respectively.}
	\label{fig:bessel-first-kind}
\end{figure}

In a physical context, Bessel functions capture the collection of vibration modes emanating from a striken drumhead \cite{strauss2008pde}. Although $J_k(x)$ can be written in an infinite series, its asymptotic form (Equation \ref{eq:FBsPCA-bessel-asymp}) better encapsulates the damped oscillation characteristics. Figure \ref{fig:bessel-first-kind} explicitly plots $J_k$ with three different $k$. Here $R_{k, q}$ is exactly the $q$-th point from the left where $J_k$ crosses the horizontal axis. In our context, our argument $R_{k, q} \frac{r}{\gamma}$ within Equation \ref{eq:FBsPCA-psi} scales the first $q$ nodes to support the Fourier disk of radius $\gamma$.
\begin{equation} \label{eq:FBsPCA-bessel-asymp}
J_k(\rho) \approx \sqrt{\frac{2}{\pi \rho}} \cos\left( \rho - \frac{\pi}{4} - \frac{n \pi}{2} \right) + O\left( \frac{1}{\rho^{3/2}} \right)
\end{equation}

Noting the fact that  since the bases $\psi^{k, q}(r, \phi)$ are truncated outside some radial length $\gamma$ in the Fourier domain, the information at frequencies higher than $\gamma$ don't contribute. One advantage to consider such band-limiting is that it reflects the Cryo-EM envelope function which dampens higher frequency coefficients; in other words, it is reasonable to neglect the Fourier coefficients on the exterior $r > \gamma$ because they are vanishing as $r$ grows larger. To summarize, the  truncated Fourier-Bessel approximation is very relevant to capturing the 2D particle projection decomposition bounded within circles, both in image space and in the  Fourier transform.


Alternatively, it  should be noted that it is thus possible to decompose 2D particle projections in image space as in Equations \ref{eq:FBsPCA-image-inv} and \ref{eq:FBsPCA-psi-inv} using the  inversion and linearity of the Fourier transform.
\begin{equation} \label{eq:FBsPCA-image-inv}
I^{(i)}(r, \phi) = \operatorname{Re} \sum_{k = -k_{\max}}^{k_{\max}} \sum_{q = 1}^{p_k} a^{(i)}_{k, q} F^{-1}(\psi^{k, q}_{\gamma})(r, \phi)
\end{equation}
\begin{equation} \label{eq:FBsPCA-psi-inv}
F^{-1}(\psi^{k, q}_{\gamma})(r, \phi) = \frac{2\gamma \sqrt{\pi} (-1)^q R_{k, q} J_k(2 \pi \gamma r)}{\iota^k ((2\pi\gamma r)^2 - R_{k, q}^2)} e^{\iota k \phi}
\end{equation}

\begin{figure}[hbt]
	\centering
	\includegraphics[width=8cm,height=6cm,keepaspectratio]{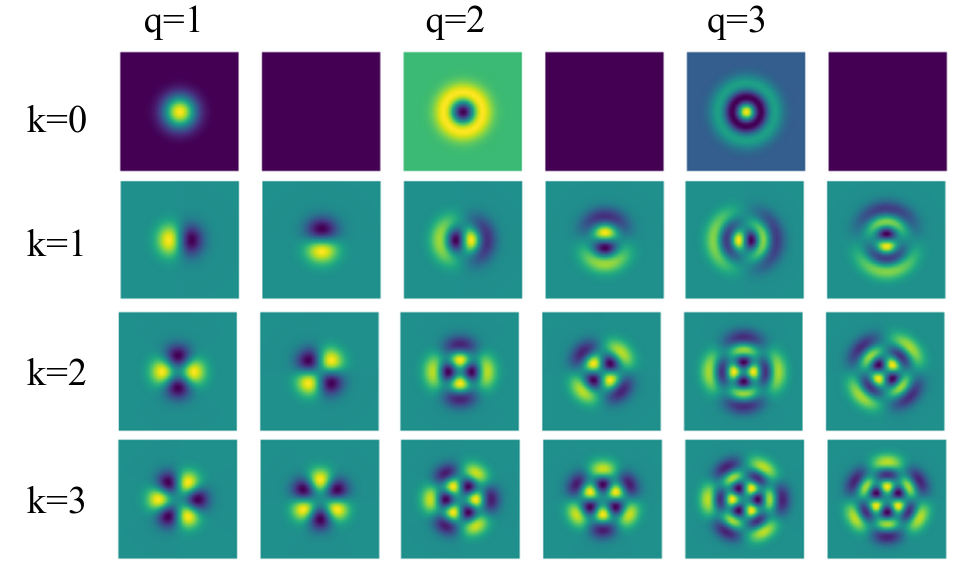}
	\caption{Examples of 12 Fourier-Bessel functions in image space $F^{-1}(\psi^{k, q})$ over frequency $k$ and $q$-th root. Odd columns show the real part while even columns show the imaginary part of the functions.}
	\label{fig:bessel_fns}
\end{figure}

Albeit, while it appears in a  messy form, Equation \ref{eq:FBsPCA-psi-inv} has a separation of variables $r$ and $\phi$ by considering it as $A \times \frac{J_k(B r)}{B^2 r^2 - C^2} \times e^{\iota k \phi}$ for constants $A$, $B$, $C$ dependent on $k$ and $q$. Figure \ref{fig:bessel_fns} shows the plots of these image-space Fourier-Bessel functions over different $k$ and $q$.

\begin{figure}
	\centering
	\includegraphics[width=0.9\linewidth,keepaspectratio]{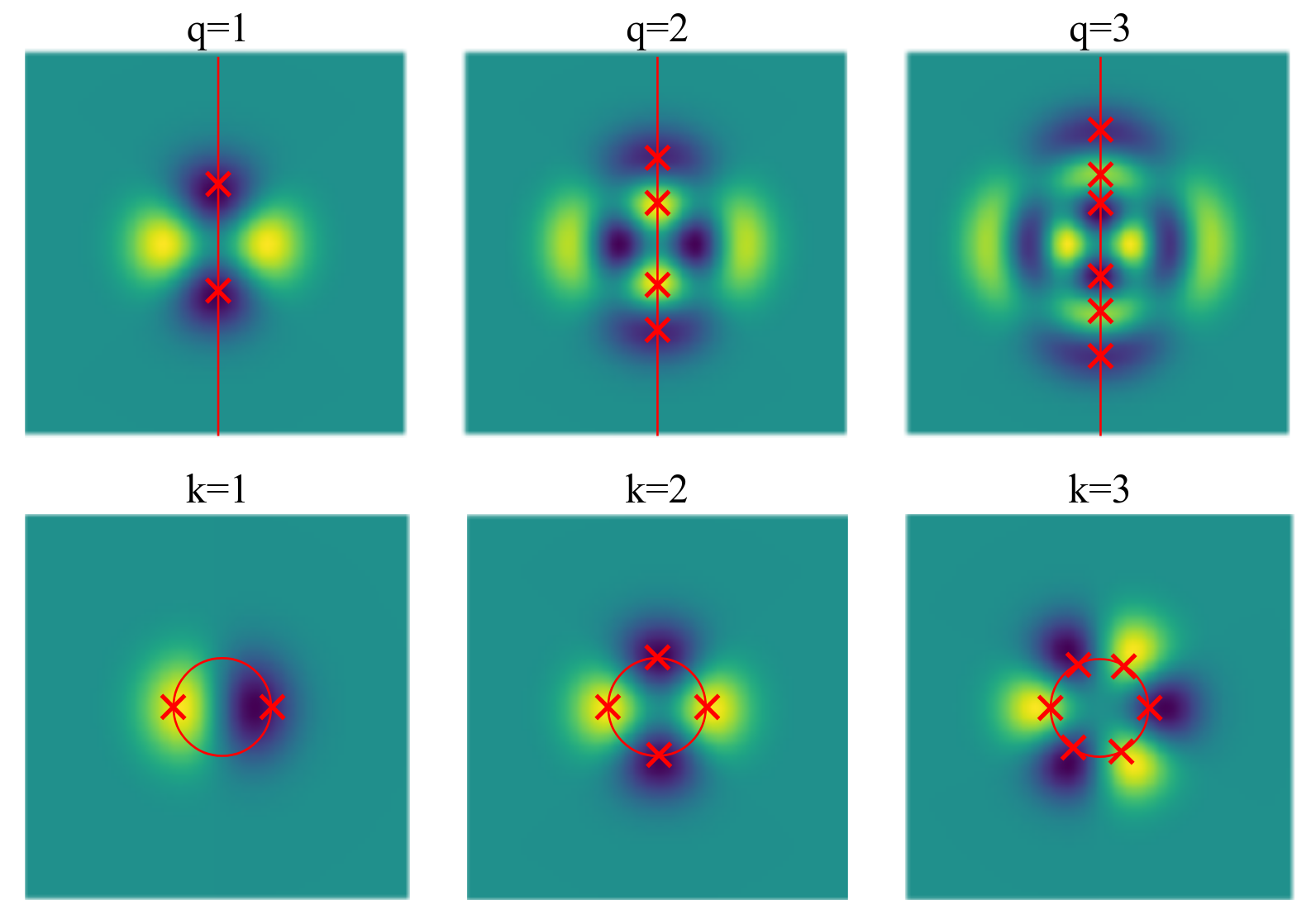}
	\caption{How to recognize $k$ and $q$ from an image expressed in the Fourier-Besseel basis: $k$ is the number of modes in an angular path while $q$ is the number of modes in a radial path.}
	\label{fig:bfns_kq}
\end{figure}

One might see the pattern of $k$ and $q$ on these Fourier-Bessel functions: $k$ relates to the number of extrema along an angular path while $q$ relates to the number of extrema along the radial path illustrated in Figure \ref{fig:bfns_kq}. This makes more intuitive sense once we look at the definitions. Note that $k$ associates with the rotational component $e^{\iota k \phi}$. On the other hand, if one increases $q$ by one, the term $J_k(R_{k, q} \frac{r}{\gamma})$ will include the next extrema of $J_k$ after the $q$-th root in the truncated disk.

\subsection{Countering Effects of Noise and Rotation}

\begin{figure*}
	\centering
	\includegraphics[height=5cm,keepaspectratio]{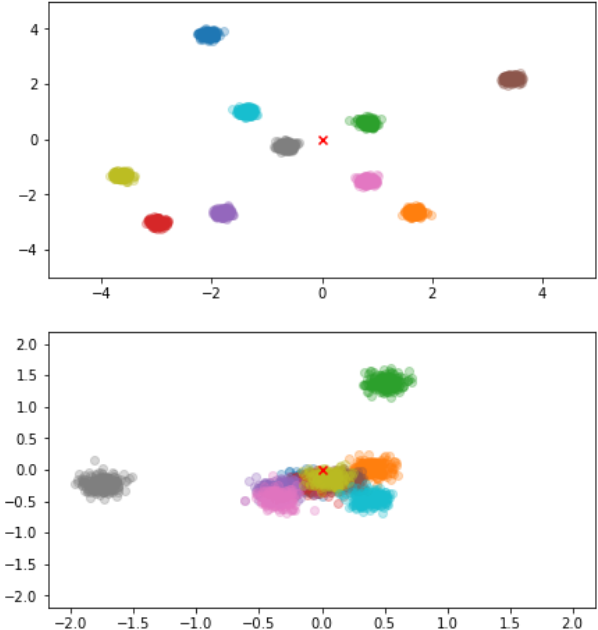}
	\includegraphics[height=5cm,keepaspectratio]{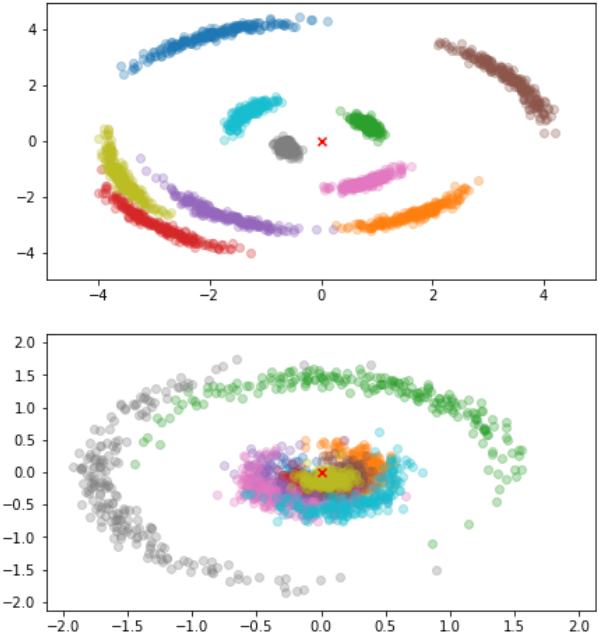}
	\includegraphics[height=5cm,keepaspectratio]{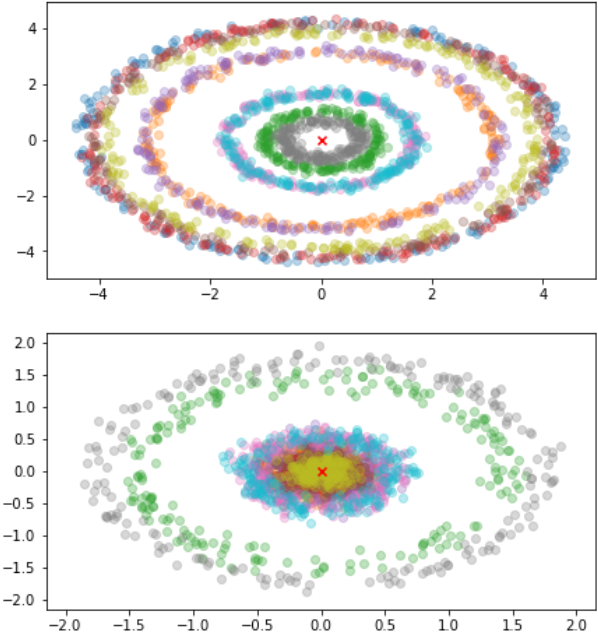}
	\caption{Samples after perturbations in data model on 2 complex FBsPCA coordinates of 10 different images. Left column: FBsPCA coordinates of images with Gaussian noise. Center column: FBsPCA coordinates of images with a combination of Gaussian noise and small planar angle deviation (standard deviation of $5$ degrees). Right column: FBsPCA coordinates of images with a combination of Gaussian noise and uniform planar angle deviation over $[-\pi, \pi]$. The angular frequencies for top and bottom rows are $\omega_1 = 2$ and $\omega_2 = 7$ respectively. The red crosses mark the origin in complex plane.}
	\label{fig:fbspca_perturb}
\end{figure*}

We sample 10 different particle images and study the variation in the  FBsPCA representation after perturbing the image with Gaussian noise and random rotations. Figure \ref{fig:fbspca_perturb} shows samples in selected FBsPCA coordinates after different perturbations. Unsurprisingly, Gaussian noise affects an image's FBsPCA representation by perturbing around the original coordinate in all dimensions. On the other hand, a random rotation within a small angle creates a band of distribution due to the steerability. Such a band shape can be also be found when the rotation alignment contains some error. Through the use of polar metrics, PolarGMM leverages these observations to model and measure noisy particle images under imperfect alignment.

In the same light, the plots on the right column portray a typical set of unaligned particle images in FBsPCA; samples form rings in all FBsPCA dimensions. In an ideal scenario, a perfect rotation alignment would neglect the rotation effect and reproduce the plots on the left column, making the samples concentrated around the FBsPCA of the average particle image.

\subsection{Effects of Translation}

\begin{figure}[H]
	\centering
	\includegraphics[width=0.45\linewidth,keepaspectratio]{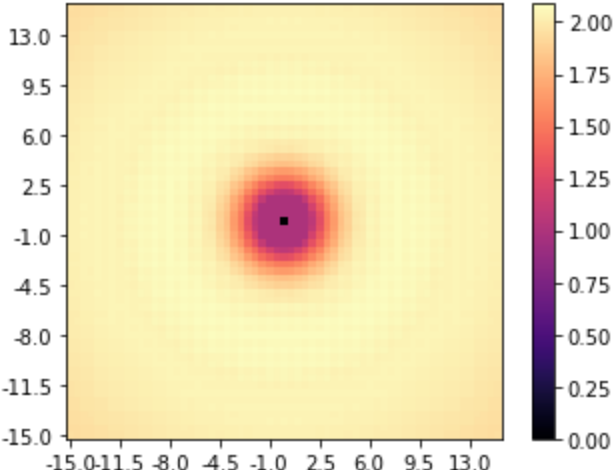}
	\includegraphics[width=0.45\linewidth,keepaspectratio]{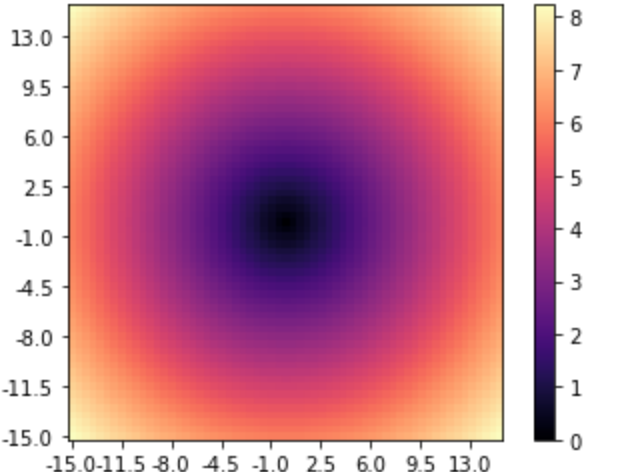}
	\caption{Heatmaps showing $\ell^2$-norms of $\Psi_\bft - I$ (left) and $\mu_\bft$ (right) as a function of $\bft \in [-15, 15]^2$ pixel units, $4L^2 = 150 \times 150$.}
	\label{fig:txy-affine}
\end{figure}

How much can $T_\bft(\bfz)$ deviate from $\bfz$ as a function of $\bft$? To avoid complicated analysis due to image translation, we instead study an upper bound of the deviations. Using the Cauchy-Schwarz inequality, the magnitude of deviation is bounded by $\norm{T_\bft(\bfz) - \bfz}_2 \leq \norm{\Psi_\bft - I}_2 \norm{\bfz}_2 + \norm{\bfmu_\bft}_2$. Figure \ref{fig:txy-affine} shows $\norm{\Psi_\bft - I}_2$ on the left and $\norm{\bfmu_\bft}_2$ on the right. Whenever the $\norm{\bft}_2$ is large, the translated FBsPCA vector is very different from $\bfz$, implying that PolarGMM would not work directly . This emphasizes the importance of our introducing an algorithmic Expectation Maximization centering step.

\begin{figure*}
	\centering
	\includegraphics[width=0.75\linewidth,keepaspectratio]{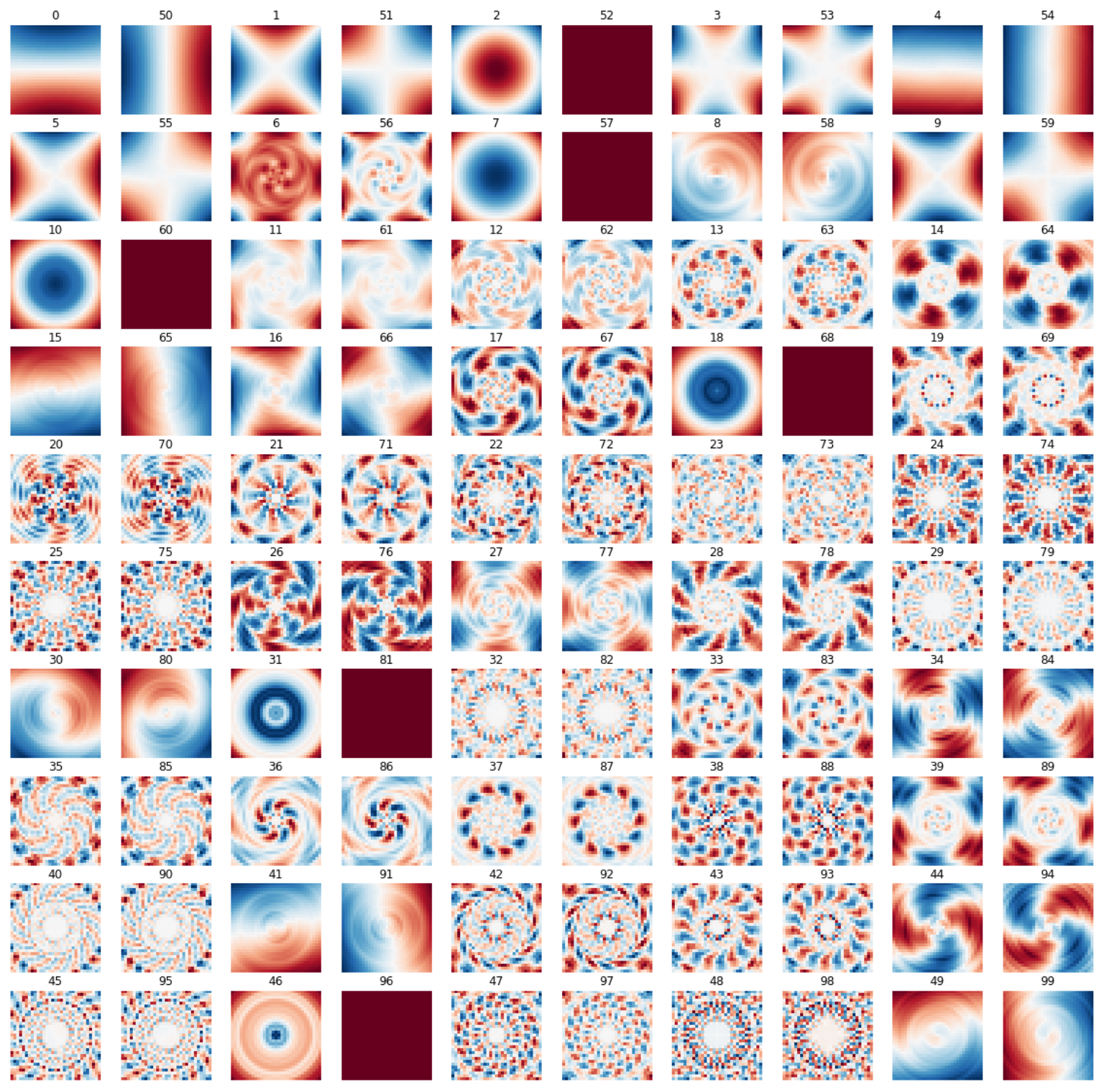}
	\caption{Heat-maps of 50 complex FBsPCA coordinates of $\bfmu_\bft$ as a function of planar translation vector $\bft$ where the center of images corresponds to $\bft = 0$. Real coordinates are labeled $0, \dots, 50$ while Imaginary ones are labeled $51, \dots, 100$. $i$-th complex dimension corresponds to two plots with labels $i$ and $i+50$. Red and blue colors respectively indicate positive and negative values of the particular coordinate of $\bfmu_\bft$.}
	\label{fig:txy-mu-xy}
\end{figure*}

As a side note, $\mu_{\bft} \in \mathbb{C}^m$ has a number of elements sufficiently small to be visualized and, in fact, possesses interesting patterns shown in Figure \ref{fig:txy-mu-xy}.

\section{PolarGMM and Alignment in Motion}

Figure \ref{fig:polargmm-convergence} visualizes the main loop of our algorithm. This includes the alignment and the PolarGMM,  EM steps. Each frame of the animation shows six plots of selected FBsPCA coordinates in complex planes. The blue dots represent individual particle image samples after alignment. The colored bands represent individual PolarGMM clusters where the arc length depicts the cluster's angular deviation $\sigma_{\phi,j}^{(c)}$ and the radial thickness depicts the cluster's radial deviation $\sigma_{r,j}^{(c)}$.

The animation displays a quick convergence in many coordinates, most visibly on the top-left plot. For some coordinate such as the top-right one, PolarGMM and alignment take a few more iterations to converge. These slower coordinate convergence correspond to higher-frequency FBsPCA bases which are more sensitive to rotation. The progress however in these coordinates indicates a finer and finer alignment.


\begin{figure*}
	\centering
	\animategraphics[loop,autoplay,width=1.0\linewidth]{12}{images/cgmm_500}{1}{9}
	\caption{Iterations of PolarGMM and alignment.}
	\label{fig:polargmm-convergence}
\end{figure*}

\section{Rotation Alignment}

\begin{figure}
	\centering
	\includegraphics[width=\linewidth,keepaspectratio]{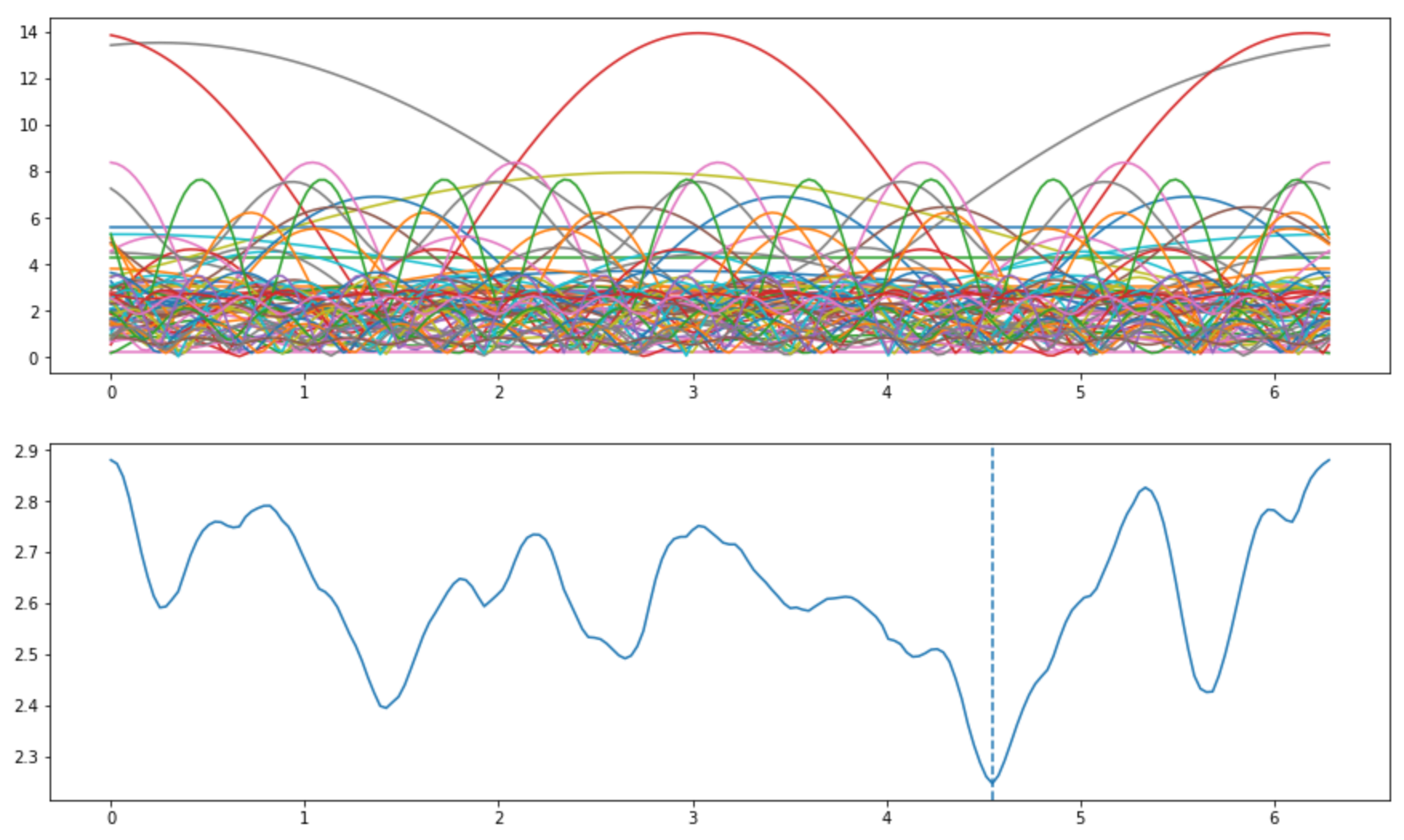}
	\caption{Distance of FBsPCA vectors as a function of planar rotation angle. Top: each line represents the complex distance of an entry in the vectors. Bottom: sum of all lines on top, i.e. the square distance. The dotted line marks the optimal alignment angle between the two vectors.}
	\label{fig:fbspca_dist}
\end{figure}

2D particle alignment is a non-trivial task, even when considering only rotation under FBsPCA. The best alignment between two vectors minimizes the distance between them. The difficulty of alignment thus associates with the complexity of the distance function in FBsPCA space. Figure \ref{fig:fbspca_dist} shows the distance of two FBsPCA vectors as a function of alignment angle $\alpha$. It illustrates how non-convex the alignment objective function can be.
 
Hence, the equally spaced angle samples of sufficiently large size $n_\alpha$ guarantee to find the optimal alignment within $\pi / n_\alpha$ radian. As opposed to iterative optimization, such strategy trivially invites parallelism to speed up the alignment search.

\section{Translation Alignment}

\begin{figure}
	\centering
	\includegraphics[width=0.9\linewidth,keepaspectratio]{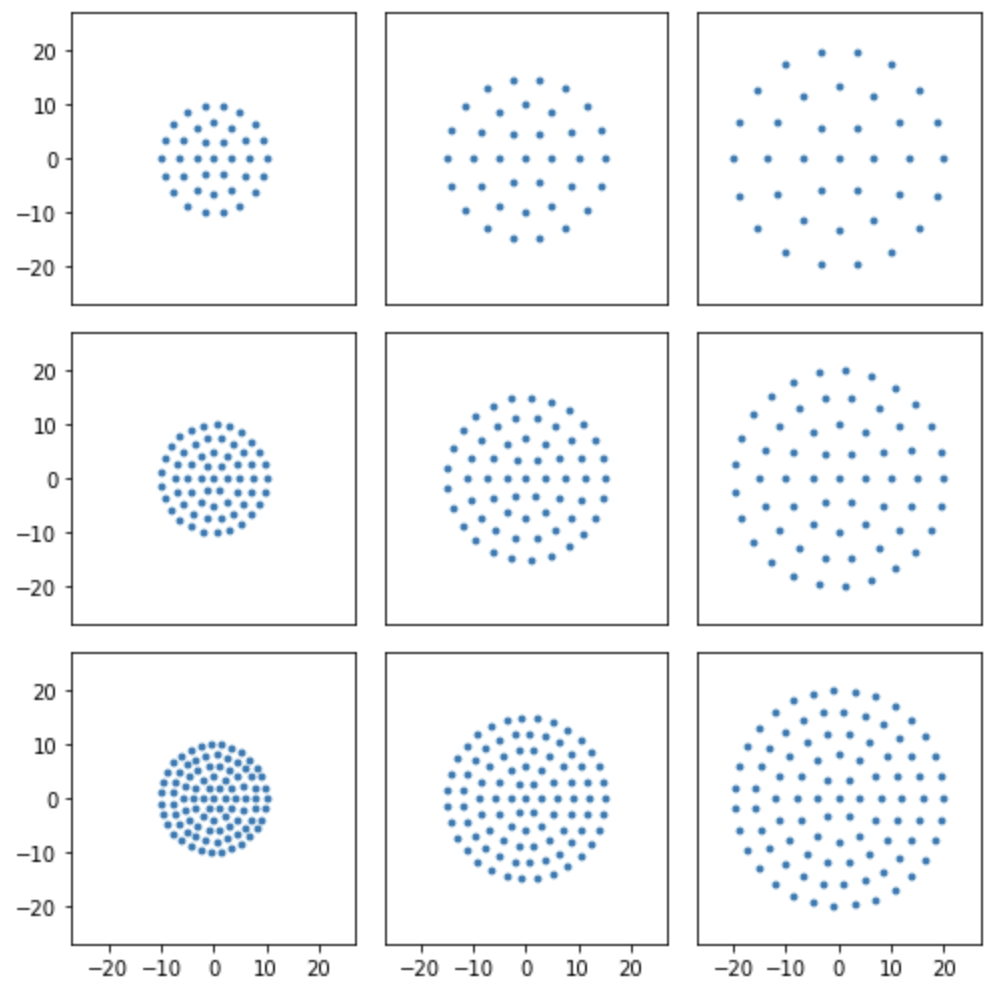}
	\caption{Translation grid consisting of rings. From left to right columns, we vary the limiting radius $R \in \{10, 15, 20\}$. From top to bottom rows, we vary the number of rings $n_r \in \{3, 4, 5\}$. Note that the positions are rounded to the nearest pixels.}
	\label{fig:txy-grid}
\end{figure}

Figure \ref{fig:txy-grid} shows different configured grids of translation samples. In the experiments, we set $n_r = 4$ rings spanning within a radius $R = 15$, producing a picture in the middle row and column of the figure.

As a reminder, we use this grid to search for translation alignment from cluster's center images to all batch samples. This is computationally cheaper than the reverse: alignment from samples to centers. Specifically, the alignment heavily uses the cached translation operator. Before searching, it pre-computes $\Psi_\bft$ and $\bfmu_\bft$ for all translations $\bft$ in the translation grid. The alignment then searches for the best translation by applying a cached translation lookup. 

Since the alignment inverts the direction of translation and rotation application, PolarGMM's EM step translates samples in the opposite direction prior to updating PolarGMM parameters. These translations contain the inverse of the rotation operator, which significantly increases the feasible set of possible translations. Caching a very large set of possible translations trades-offs precomputation time, intermediate memory utilization with  alignment speed. Hence we additionally provide PolarGMM updates that can utilize a vanilla translation operator.

\end{document}